\documentclass[lettersize,journal]{IEEEtran}
\usepackage[numbers]{natbib}
\usepackage{amsmath,amsfonts}
\usepackage{algorithmic}
\usepackage{algorithm}
\usepackage{array}
\usepackage[caption=false,font=normalsize,labelfont=sf,textfont=sf]{subfig}
\usepackage{textcomp}
\usepackage{stfloats}
\usepackage{url}
\usepackage{verbatim}
\usepackage{graphicx}
\usepackage{cite}
\usepackage[nolist]{acronym} 
\usepackage{booktabs}
\usepackage{multirow}
\usepackage{amssymb}
\usepackage{xcolor}
\usepackage[dvipsnames]{xcolor}
\usepackage{siunitx}
\usepackage{bbm}
\usepackage{microtype}

\usepackage[pagebackref=true,breaklinks=true]{hyperref}
\hypersetup{
     colorlinks,
     linkcolor={red!75!black},
     citecolor={blue!75!black},
     urlcolor={blue!75!black},
}
\usepackage[capitalise]{cleveref}

\begin{acronym}
\acro{mae}[MAE]{Mean Absolute Error}
\acro{mme}[MME]{Mean Meter Error}
\acro{map}[AP]{Average Precision}
\acro{map1}[AP-1\%]{Average Precision - 1\%}
\acro{map5}[AP-5\%]{Average Precision - 5\%}
\acro{ViT}[ViT]{Vision Transformer}
\acro{CNN}[CNN]{Convolutional Neural Network}
\acro{aspp}[ASPP]{Atrous Spatial Pyramid Pooling}
\acro{gpr}[GPR]{Ground-Penetrating Radar}
\acro{dl}[DL]{Deep Learning}
\acro{tn}[TN]{True Negative}
\acro{fp}[FP]{False Positive}
\acro{fn}[FN]{False Negative}
\acro{tp}[TP]{True Positive}
\acro{gru}[GRU]{gated recurrent unit}
\acro{sgd}[SGD]{stochastic gradient descent}
\acro{mlp}[MLP]{multilayer perceptron}
\acro{geoconv}[GeoConv]{geographically aware convolutional module}
\end{acronym}

\begin{document}

\title{ICIBS: Increasing the Context for Ice Boundary Segmentation}

\author{Marcel Dreier$^{1, *}$, Dakota Pyles$^2$, Nora Gourmelon$^1$, Moritz Koch$^2$, Fei Wu$^1$, Thorsten Seehaus$^2$, Matthias Braun$^2$, \\Andreas Maier$^1$, Vincent Christlein$^1$
\thanks{$^*$ Corresponding author, marcel.dreier@fau.de\\
$^1$Pattern Recognition Lab, Computer Science Department,
Friedrich-Alexander-Universität Erlangen-Nürnberg, Erlangen, Germany\\
$^2$Institut für Geographie, Department of Geography and Geosciences,
Friedrich-Alexander-Universität Erlangen-Nürnberg, Erlangen, Germany\\
}%
\thanks{This work has been submitted to the IEEE for possible publication.
Copyright may be transferred without notice, after which this version may
no longer be accessible.}}

\markboth{Preprint submitted to IEEE Transactions on Geoscience and Remote Sensing}%
{Dreier \MakeLowercase{\textit{et al.}}: ICIBS: Increasing the Context for Ice Boundary Segmentation}

\maketitle
\bstctlcite{IEEEexample:BSTcontrol}
\begin{abstract}
Mountain glaciers and ice sheets are undergoing rapid loss of ice mass. To assess their current volume and ice-thickness distribution, and to accurately forecast their future, ground-penetrating radar systems are employed to measure ice thickness. These instruments produce radio-echo sounding data that reveal the internal structure of the ice and its boundaries at the surface and beneath the ice body. However, delineating these interfaces is time-consuming as the surveyed area can span several hundred kilometers. To accelerate the labelling process, automated methods have emerged, but they often struggle to delineate the ice bottom, particularly in temperate glaciers. The high water content inherent in temperate ice increases both scattering and attenuation of the radar signal, resulting in weak and ambiguous reflections at the ice bottom and generally noisier data. As a result, the visual features corresponding to the ice bottom become ambiguous. To address this issue, we propose several extensions to deep learning models that incorporate additional contextual information into the prediction. Specifically, we integrate modifications that incorporate additional spatial and geo-spatial contextual information into the network. With these enhancements, we achieve a new state-of-the-art performance on the IceAnatomy benchmark dataset, with mean errors of 10.9\,m for the FAU subset and 43.4\,m for the CReSIS subset in a single combined system. We also demonstrate that our method generalises well to previously unseen data, outperforming the baseline by 49.6\% on new campaign data from Patagonia. The model code and the new dataset are available at \url{https://github.com/ki7077/ICIBS/tree/main} and \url{https://zenodo.org/records/23392580}.

\end{abstract}

\begin{IEEEkeywords}
remote sensing, GPR, deep learning, radio echo-sounding.
\end{IEEEkeywords}

\section{Introduction}
\label{sec:intro}

Estimates indicate that 49--83\,\% of mountain glaciers will disappear by 2100 \citep{rounce2023global}, which will have a strong impact on local environments, ecosystems, and communities~\citep{Intergovernmental_Panel_on_Climate_Change_IPCC2023-gl,IPCC6}. 
These projections  highlight the need for robust baseline assessments of glacier thicknesses at specific points in time. However, among the more than 200,000 glaciers located outside Greenland and the Antarctic Ice Sheet, less than 5,000 have concrete ice thickness measurements~\citep{Hock_Maussion_Marzeion_Nowicki_2023,essd-12-3039-2020}. To address this data scarcity, scientists conduct field campaigns using ground-penetrating radar systems to measure ice thickness in previously unobserved glaciers~\citep{koch2025state}. The systems operate by transmitting electromagnetic waves into the ice and recording the returning signals, producing radargrams or image-like maps of subsurface reflection amplitudes via inversion techniques~\citep{blindow2011performance,allen2012antarctic}.
These radargrams enable the interpretation of internal glacier structures along with the delineation of ice boundaries, respectively, the ice thickness, which are essential variables for predicting the future glacier evolution. However, radar surveys accumulate vast amounts of data, making manual analysis increasingly impractical~\citep{sime2011automated}.

To address this issue, algorithms and deep learning models have been developed to automatically delineate the ice boundaries~\citep{tc-19-2159-2025}. These approaches have achieved promising results, particularly for cold glaciers, where the recorded radar reflection amplitudes remain largely undistorted. In contrast, the results for temperate and polythermal glaciers are often worse, as water attenuates and scatters the signal, complicating echo interpretation~\citep{me}. Likewise, thick glaciers can attenuate the received signal, further complicating the distinction between internal features. In such cases, accurate delineation of the ice bottom requires frequent manual interpretation. However, that raises the question of why human annotators are better at interpreting radar data. 

\begin{figure}[t]
\centering
\includegraphics[width=0.45\textwidth]{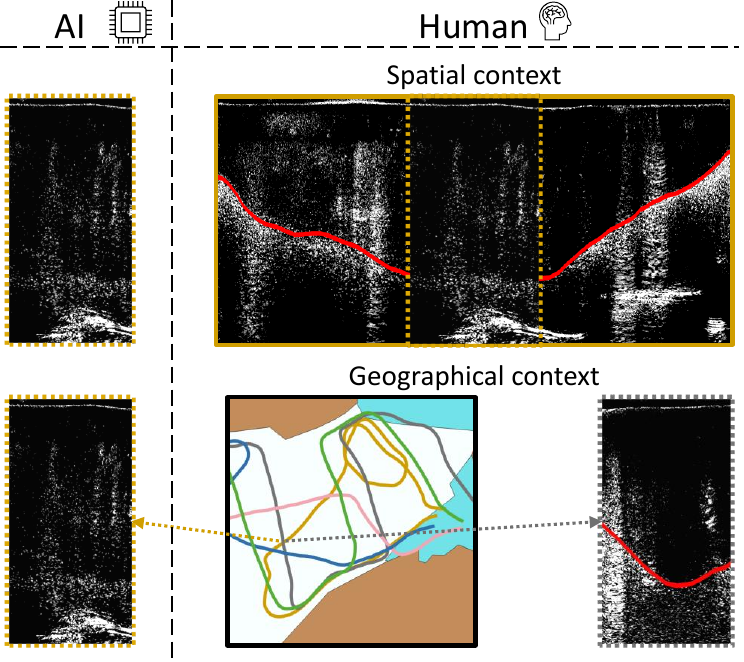}
					\caption{Comparing how deep learning models commonly view the problem in contrast to humans: Deep learning models typically approach radargram interpretation using a restricted spatial context to reduce computational complexity, relying primarily on local visual features for boundary prediction. In contrast, human annotators interpret radar data within a substantially broader spatial framework and additionally incorporate information from geospatially proximate radargrams to guide their decisions. The corresponding ice bottom is marked in \fboxsep=1pt\colorbox{red!100}{red}. The borders of the radargrams are color-coded according to their corresponding flight in 
                    \fboxsep=1pt\colorbox{gray!100}{\textcolor{white}{gray}} and \fboxsep=1pt\colorbox{Dandelion!100}{dark yellow}.}
\label{fig:intro}
\end{figure}

When comparing the processing of automatic approaches and humans in Fig.~\ref{fig:intro}, we can see that automatic methods commonly process radargram transects in a patchwise manner to stabilize model training and to limit computational costs~\citep{dong2021eisnet,cai2022accurate,kamangir2018deep,liu2022ice,jebeli2023tssa,me}. While this strategy achieves the intended goals, it also restricts the model's ability to explore the full spatial context, unlike a human who can view a larger section of the radargram. Furthermore, human annotators are not limited to the spatial features of the radargram when predicting ice boundaries. Instead, they can examine the surveyed area and flightlines, cross-reference them with radargrams from nearby regions, and use this information to infer the glacier’s internal structure. In ambiguous cases, they may also consult other experts to come to a conclusion. 

Motivated by these observations, we aim to incorporate such human-inspired contextual reasoning into a deep learning framework for automatic ice-boundary delineation. Specifically, we propose:

\begin{itemize}
    \item A novel approach to incorporate the local geospatial context;
    \item A new hooking mechanism for ice bottom detection in radio echo-sounding data;
    \item A new state-of-the-art ensembling model to refine the predictions.  
\end{itemize}\

To provide an overview of the general topic, we first review existing work on ice boundary extraction from radio-echo sounding data in Section \ref{sec:relatedwork}. In Section \ref{sec:methodology}, we present our proposed modifications and demonstrate how each can be integrated into a simple U-Net architecture. To empirically validate the efficacy of our approach, we evaluate our model on the IceAnatomy benchmark dataset in Section \ref{sec:exp} and then discuss potential weaknesses and directions for future work in Section \ref{sec:discussion}. Lastly, we conclude our work in Section \ref{sec:conclusion} by summarizing our results and contributions.

\section{Related Work}
\label{sec:relatedwork}
Radargrams of ice enable the extraction of a wide range of subsurface features, including snow layers, shallow stratigraphy, and ice layers~\citep{tc-19-2159-2025}. This study, however, focuses specifically on the automatic delineation of the ice bottom. To solve this task, prior work has proposed a wide variety of methods, including contour-based approaches~\citep{gifford2010automated,Mitchell.2013,Rahnemoonfar.2017b,Rahnemoonfar.2017}, graphical models ~\citep{Crandall.2012,Berger.2018,Xu.2017,Lee.2014}, classical machine learning approaches \citep{Ilisei.2014,Ilisei.2015}, and deep learning models~\citep{tc-19-2159-2025}.

Deep learning networks have become increasingly prevalent due to their ability to capture complex spatial patterns and their flexibility in adapting to the often noisy data distribution. These models are typically designed either to directly predict ice boundaries \citep{dong2021eisnet,cai2022accurate,kamangir2018deep,liu2022ice,jebeli2023tssa,me} or to segment the input radargram into distinct semantical regions from which interfaces can subsequently be derived \citep{donini2021deep,garcia2021automatic,ghosh2022hybrid,ghosh2022transsounder,ghosh2023enhanced,ghosh2023unsupervised,ghosh2024urs}. Among the various architectures, the convolutional U-Net \citep{dong2021eisnet, jebeli2023tssa, me, donini2021deep, ghosh2022transsounder, ronneberger2015u} remains the most widely adopted, as it effectively combines multi-scale semantic information with fine-grained spatial detail. Nevertheless, alternative architectures—including ResNet-based models~\citep{cai2020end}, hybrid convolutional neural networks~\citep{ghosh2022hybrid,ghosh2022transsounder}, Vision Transformers~\citep{ghosh2023enhanced,ghosh2023unsupervised,ghosh2024urs}, and convolutional encoders~\citep{cai2022accurate,kamangir2018deep,liu2022ice,garcia2021automatic}, have also been explored. 

A fundamental limitation of deep learning models is their limited spatial view of the input. Expanding the spatial resolution of the input typically increases the required memory and computational demands quadratically. Moreover, expanding the spatial dimensions often requires deeper and heavier network architectures, which further amplifies model complexity and resource consumption. Thus, larger input data is commonly processed using a sliding-window approach, sacrificing spatial context to remain computationally efficient.

Rijthoven et al.~\citep{VANRIJTHOVEN2021101890} aimed to address this by proposing Hooknet, a dual-branch architecture comprising a target branch and a context branch. The target branch processes the high-resolution input patch, while the context branch captures a broader spatial region via a downsampled representation centred on the same location. Information exchange between the branches is achieved by concatenating the context branch feature maps with those of the target branch at the bottleneck layer of the target branch, thereby enabling the integration of broader contextual information without substantially increasing the depth of the target network.

The framework is inherently extensible: multiple context branches can be incorporated to further enlarge the spatial field of view~\citep{sneayknet}, and enhanced coupling strategies have been developed to promote multi-resolution feature exchange~\citep{sneayknet,amdhooknet}. Although hooking-based architectures have achieved promising results in the medical domain~\citep{VANRIJTHOVEN2021101890,sneayknet,van2024multi}, document analysis~\citep{wu2025lightweight}, and remote sensing~\citep{amdhooknet,wu2024contextual,11296938}, their potential has not been explored for radio-echo sounding data yet.

\section{Methodology}
\label{sec:methodology}
In this section, we explain the structure of our modifications. We start with the U-Net baseline %
and successively add our novel hooking mechanism and geography-aware convolutions.

\begin{figure*}[htbp]
\centering
\includegraphics[width=0.95\textwidth]{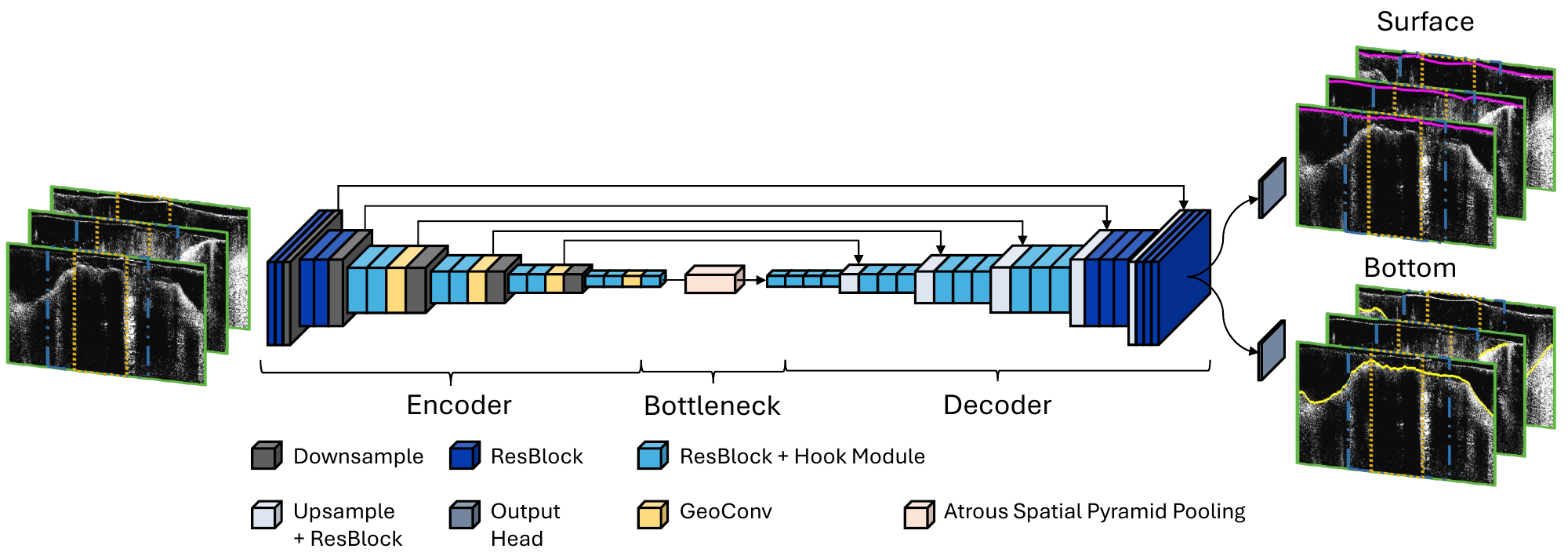}
					\caption{Overview of our proposed architecture, adapted from Dreier et al.~\citep{me}.
                    The model receives three geospatially proximate radargram sections, from which three context windows are cropped, as indicated by the outline (\fboxsep=1pt\colorbox{Dandelion!100}{small}, \fboxsep=1pt\colorbox{NavyBlue!100}{\textcolor{white}{medium}}, and \fboxsep=1pt\colorbox{LimeGreen!100}{large}). To enable joint processing, the nine radargrams are normalized to a consistent size. For each radargram, the network predicts both the \fboxsep=1pt\colorbox{Rhodamine!100}{ice surface}
                    and the \fboxsep=1pt\colorbox{Yellow!100}{ice bottom} interfaces. While both outputs are generated during training, the ice surface prediction task serves exclusively as an auxiliary task and is not considered for the evaluation.}
\label{fig:fullmodel}
\end{figure*}

\subsection{Base Model}
\label{meth:base}
As a baseline, we employ a U-Net architecture following the design by Dreier et al.~\citep{me}, which consists of an encoder, a bottleneck, and a decoder. The encoder comprises five stages, each containing two residual blocks and a downsampling block that halves the spatial resolution of the feature map. In the bottleneck, the network combines residual blocks with atrous spatial pyramid pooling~\citep{aspp} to capture information from multiple effective receptive fields. The decoder mirrors the encoder structure but increases the number of residual blocks to three per stage and replaces the downsample blocks with upsample blocks that progressively restore the original spatial resolution. Additionally, skip connections between the encoder and the decoder merge high-resolution spatial features with deep semantic information. The final predictions for the ice surface and ice bottom are produced by two task-specific heads. Training relies on the combined regression and classification loss proposed in Dreier et al.~\citep{me}, with a gradual linear increase of the regression loss contribution during the initial 200 training steps to stabilize optimization.

To strengthen the base model's contextual awareness, we incorporate into our proposed model two additional components: a hooking module and a \ac{geoconv}. The hooking module facilitates cross-resolution information exchange to enlarge the effective spatial context and is appended to the end of the residual block. The \ac{geoconv} captures geospatial context by aggregating information from three geospatially adjacent radargrams, providing the network with an extended geospatial perspective. To limit the computational overhead, the hooking module is applied only to residual blocks below the second level, while \ac{geoconv} is restricted to the encoder. Figure~\ref{fig:fullmodel} presents an overview of the model structure. 

\subsection{Geographical Sampling Strategy}
\begin{figure}[htbp]
\centering
\includegraphics[width=0.45\textwidth]{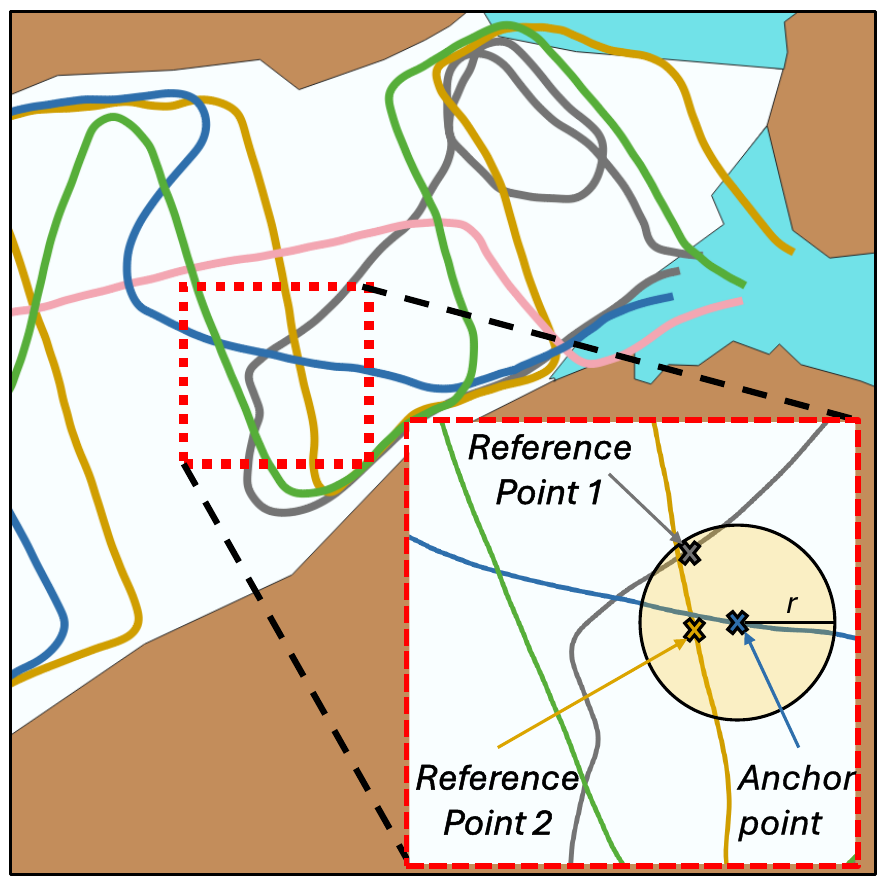}
					\caption{Illustration of the proposed sampling strategy for geospatially adjacent radargrams using flight lines acquired over the Perito Moreno Glacier. The background schematically represents the glacier environment, distinguishing \framebox{glacier ice}, \fboxsep=1pt\colorbox{Tan!100}{\textcolor{black}{rock}}, and \fboxsep=1pt\colorbox{SkyBlue!100}{water}. Individual flight lines are shown in \fboxsep=1pt\colorbox{LimeGreen!100}{green}, \fboxsep=1pt\colorbox{gray!100}{\textcolor{white}{gray}}, \fboxsep=1pt\colorbox{NavyBlue!100}{\textcolor{white}{blue}}, \fboxsep=1pt\colorbox{pink!100}{pink}, and \fboxsep=1pt\colorbox{Dandelion!100}{dark yellow}. The sampling procedure first selects an anchor point and then samples two additional reference points within a predefined radius $r$. Candidates from nearby, distinct flight lines are prioritized to promote diversity among the sampled radargrams. If no suitable candidates are available within this neighborhood, the procedure falls back to sampling more distant points along the same flight line.}
                   
\label{fig:geoSampling}
\end{figure}

Identifying geospatially proximate radargram sections for a given target radargram is essential for providing rich and diverse contextual information. Choosing sections from the same flight and close to the target in sequential recording order may introduce redundant information, whereas sampling geospatially distant transits weakens the underlying correlation. To balance these competing factors, we propose a three-step sampling strategy, as illustrated in Fig.~\ref{fig:geoSampling}.

First, the midpoint of the target radargram is defined as an anchor point. Second, around this anchor, a circular region with radius $r$ is constructed, containing all potential reference points. Third, within this region, two reference points are iteratively selected along available flight paths.

The selection process follows two rules. Preference is given to geographically close sections from previously unselected flights, thus maximizing spatial correlation while preserving diversity. If no such candidates are available, the selected point maximizes its distance from already chosen references of the same flight to reduce redundancy. To increase variability among training samples, these two rules of the selection process are relaxed during training. Specifically, preference is only given to samples from distinct flights, and the distance to the anchor point is ignored.

\subsection{Geospatial Context}
\label{meth:geocontext}
To incorporate the geospatial context of the sampled radargrams into the model, it is first necessary to analyze the spatial structure of the data.  

Radio-echo sounding data are acquired using a moving radar system, commonly during airborne surveys. Electromagnetic waves are emitted from a transmitter, and the reflected signals are digitally recorded and stacked to form a trace representing the vertically stacked backscatter at a single point. Each trace is  assigned to a unique trace number indicating its position within the sequence of observations along the flight path and georeferenced. Horizontally stacked traces from continuous measurements are then displayed as a radargram (or B-scan), which shows the relationship between neighbouring traces. However, this representation considers only the sequential ordering of measurements and ignores the underlying geospatial structure. Consequently, traces that are geographically proximate but acquired during different flights or at substantially later times within the same flight are not positioned closely in the radargram, preventing vision-based models from effectively inferring their connection.
To address this issue, we introduce the \ac{geoconv} module, which is specifically designed to combine geospatially proximate features.

\begin{figure}[t]
\centering
\includegraphics[width=0.35\textwidth]{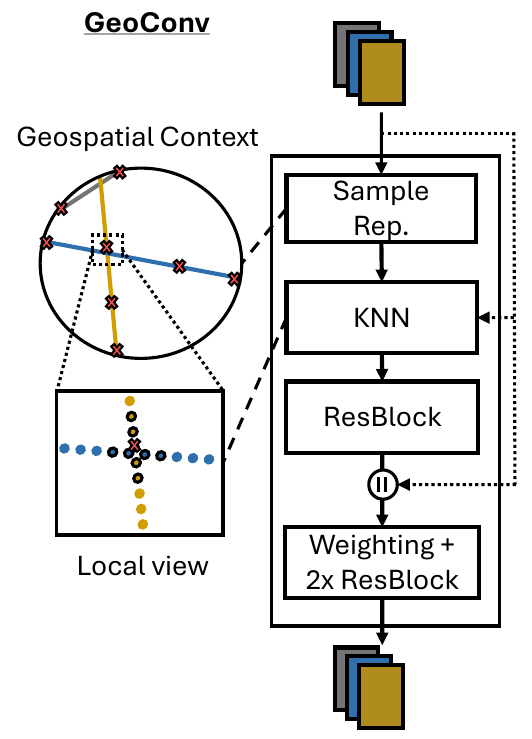}
					\caption{\ac{geoconv} module: The inputs are three feature maps. The module first iteratively samples eight traces that maximize the geographical distance from the previously sampled traces (marked with \textcolor{red}{X}). For each sampled trace, the eight nearest traces (outlined in black) are then sampled to approximate the local region. A ResBlock aggregates the local information into the eight initial traces. Finally, two additional residual blocks propagate the aggregated information to every input trace, weighting the eight sampled traces according to their geographical distance from the corresponding input trace.}
\label{fig:gie}
\end{figure}
\ac{geoconv} is based on the idea of processing radar data as point clouds rather than conventional visual features. It views radargrams not as two-dimensional images; instead, radargrams are seen as a set of georeferenced traces. Each trace is treated as a point with an x- and y-coordinate corresponding to its geospatial location. This representation enables the use of neural network layers specifically designed for point cloud analysis. In particular, \ac{geoconv} builds upon the PointConv framework~\citep{pointconv}. 

\ac{geoconv} receives three feature maps derived from geographically close radargram segments, forming a set of geospatially proximate traces $C$. To capture the global structure of the covered region, \ac{geoconv} first samples eight representative traces from $C$. These traces are selected iteratively such that each newly chosen trace maximizes its geographical distance from the previously selected ones until the eight traces are selected, forming a broad and coarse representation of the geospatial context. To refine this representation with more geospatial information, the k-nearest neighbors algorithm (KNN) identifies the eight closest traces for each sampled trace. The resulting neighboring traces were then horizontally concatenated and fused with the corresponding trace using a one-dimensional residual block. The resulting fused representations are then concatenated with every trace in the radargrams of the geospatial context $C$, thereby providing the entire context with geospatial information. Since the eight representatives have different proximities to every trace in the context $C$, their contributions are weighted using inverted and normalized geographical distances. Finally, two residual blocks merge the weighted representations into a single trace, restoring the original input dimensionality. An illustration of the structure is depicted in Fig.~\ref{fig:gie}.

\subsection{Hooking}
\label{meth:hook}
\begin{figure}[t]
\centering
\includegraphics[width=0.45\textwidth]{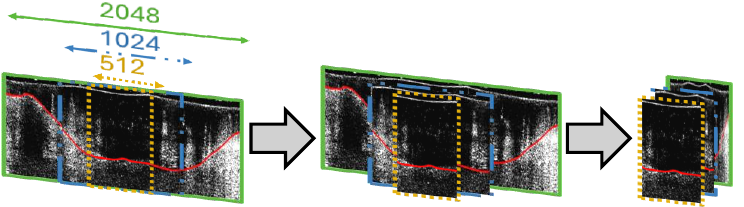}
					\caption{Preprocessing of the spatial context. An initial context patch of size $2048 \times 1024$ pixels is sampled from the dataset. Two smaller subregions of size $1024 \times 1024$ and $512 \times 1024$ pixels are then extracted from its center. Finally, all three patches are resized to $512 \times 1024$ pixels to ensure a consistent input format.}
\label{fig:preprocesshook}
\end{figure}

Building upon the approach of Rijthoven et al.~\citep{VANRIJTHOVEN2021101890}, we propose a specialized hooking component tailored to radargram data. Highly detailed inputs are essential for precise ice-boundary delineation, whereas downsampled representations with a bigger field of view provide more context for identifying deeper boundaries. To integrate these complementary scales, we sample three vertically centered windows from one radargram at small ($512\times1024$), medium ($1024\times1024$), and large ($2048\times1024$) resolutions. The small resolution is the target resolution, while the medium and large resolutions serve as additional context. To maintain a consistent and efficient input size for the network, all patches are rescaled to a uniform input size of $512\times1024$, corresponding to fine-, medium-, and coarse-resolution inputs.
Figure~\ref{fig:preprocesshook} illustrates the preparation of the data.

In contrast to conventional hooking architectures that rely on separate branches for each resolution, we employ a single shared branch. This design choice is motivated by the natural variation in horizontal resolutions of radargrams, which stems from different acquisition campaigns, processing pipelines, and radar systems. Thus, it is necessary for a generalizable model to learn resolution-invariant features, thereby making resolution-specialized branches redundant. Additionally, it also reduces the overall parameter count. 
\begin{figure*}[htbp]
					\centering
                    \includegraphics[width=0.9\textwidth]{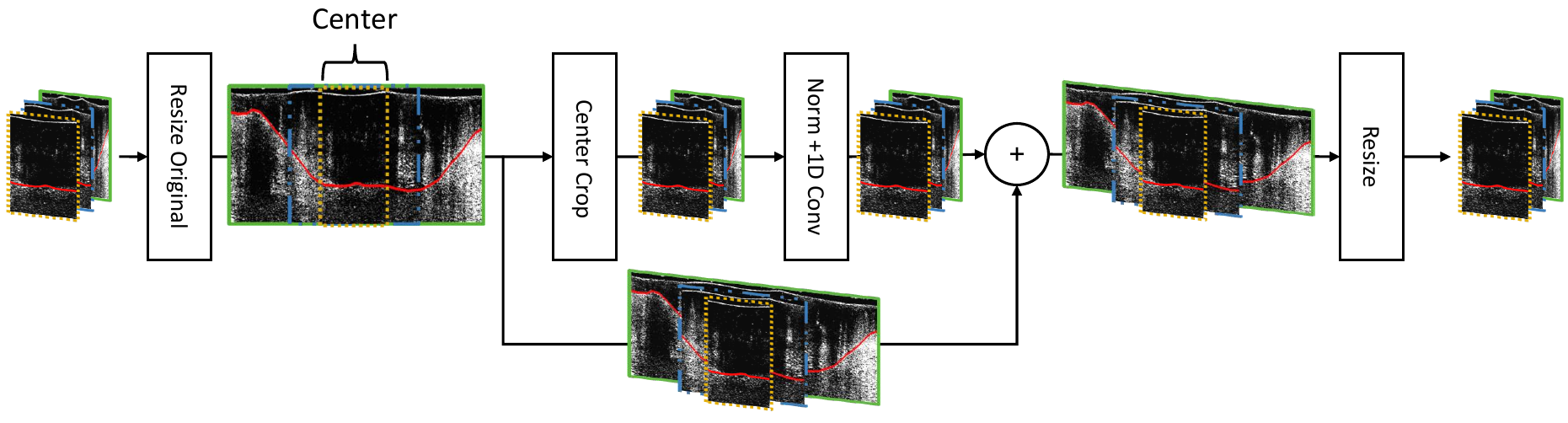}
					\caption{Illustration of the proposed hooking module. The module takes as input three vertically centered radargrams with identical dimensions but different granularity levels. First, the corresponding feature maps are rescaled to match the highest granularity among the three inputs. The central regions of the two larger rescaled feature maps are then cropped to match the spatial extent of the finest-resolution feature map. Next, the aligned feature maps are normalized and processed with a one-dimensional convolution to exchange information across the different context windows. The resulting feature representation is subsequently merged with the uncropped feature maps, and the outputs are reshaped to the original input dimensions for further processing by the network.}

\label{fig:hookingBlock}
\end{figure*}

Communication between the different context windows is realized through a dedicated hooking module as illustrated in Fig.~\ref{fig:hookingBlock}. The hooking module first adjusts the input feature maps to match the finest granularity among the three, then extracts the central region of each map aligned with the spatial extent of the smallest context window. The cropped features are normalized and combined using a zero-initialized one-dimensional convolution across the different resolutions. The processed representations are then summed up with the uncropped feature maps through a residual connection and finally rescaled to the input size. 

To enable consistent multi-resolution feature integration, the hooking module is incorporated at multiple stages of the network hierarchy. It can also be used alongside the \ac{geoconv}; however, in this configuration, each spatial location is treated as an individual sample with its own three granularity levels. Similarly, the \ac{geoconv} receives three spatially proximate radargrams for each granularity and processes each granularity independently. Since this processing strategy introduces additional computational overhead, Appendix~\ref{appendix:runtime} proposes several approaches to mitigate the associated cost.

\subsection{Ensembling}
Delineating ice boundaries in a given radargram is a complex and error-prone task, as the basal interface is often ambiguous or barely visible. When in doubt, human annotators can ask a colleague or other domain experts for help. Some studies even plan to employ multiple annotators to obtain a more reliable ground truth~\citep{multi-annotatorGPR}. We can emulate this strategy in a deep learning setting by constructing an ensemble of our proposed architecture. Specifically, we train five instances of the model from different random initializations. Each instance serves as an independent annotator that learns to predict the ice boundaries. During inference, the predicted probability maps are averaged to generate a combined prediction, thereby increasing the robustness of the final output.

\section{Experiments}
\label{sec:exp}

\subsection{Metrics}
We assess the performance of our proposed modifications with the metrics employed by Dreier et al.~\citep{me}. Specifically, we employ the \ac{mae} to quantify the absolute pixel-wise distance between the ground truth $y$ and the predicted boundary $\hat{y}$~\citep{Crandall.2012, Lee.2014, Rahnemoonfar.2017, Berger.2018, Mitchell.2013, Xu.2017, gifford2010automated, dong2021eisnet, liu2022ice}. However, since the \ac{mae} only measures the distance in pixels, the real-world impact of the application is hardly tangible. Hence, we also compute the \ac{mme}~\citep{me}, which approximates the average distance between the ground truth and the prediction in meters using the medium-dependent wave velocity $c_{\textrm{medium}}$ and the radargram-specific vertical resolution of the two-way travel time $\textrm{VR}$. The precise formulas for both are given in equations \ref{mae_equation} and \ref{mme_equation}.

\begin{align}
    \label{mae_equation}
        \text{MAE}(\hat{y},y) &=& \frac{1}{|\hat{Y}|}
        \sum_{\hat{y} \in \hat{Y}} \lvert\hat{y} - y\rvert \\
        \label{mme_equation}
       \text{MME}(\hat{y},y)  &=& \frac{c_\text{medium}}{2|\hat{Y}|} \sum_{\hat{y} \in \hat{Y}} \text{VR} \cdot \lvert\hat{y} - y\rvert
\end{align}

One downside of the distance-based metrics, like the \ac{mae} and \ac{mme}, is their sensitivity to outliers. A few wrong predictions far from the ground truth can disproportionately increase them, even when most predictions are accurate. To mitigate the effects of such a scenario, we also compute the \ac{map}~\citep{Xu.2017, gifford2010automated, liu2022ice}. This metric measures the relative frequency of correct predictions compared to the total number of predictions. As ice boundaries are usually ambiguous on the pixel level, we relax the strict correctness criterion by employing the \ac{map1} and \ac{map5}. These metrics consider predictions within a defined error bound around the ground truth label of the radargram as correct. We define their error bounds as 1\% and 5\% of the pixel height $H$ of the radargram, respectively. Equations \ref{eq:map1} and \ref{eq:map5} provide their precise mathematical definitions.

\begin{align}
    \label{eq:map1}
        \text{AP}_{1\%}(\hat{y},y) &=& \frac{1}{|\hat{Y}|}
        \sum_{\hat{y} \in \hat{Y}} \mathbbm{1}\bigg[|\hat{y} - y| < \frac{\text{H}}{100}\bigg] \\
        \label{eq:map5}
       \text{AP}_{5\%}(\hat{y},y) &=& \frac{1}{|\hat{Y}|}
        \sum_{\hat{y} \in \hat{Y}} \mathbbm{1}\bigg[ |\hat{y} - y| < \frac{\text{H}}{20}\bigg]
\end{align}

\subsection{Dataset}
\label{sec:dataset}
We train and evaluate our model on the IceAnatomy benchmark dataset~\citep{me} and its standardized training and evaluation splits. IceAnatomy consists of three distinct subsets: CReSIS, FAU, and AWI. Each one is gathered by a different institution using a different radar system, processing scheme, and focusing on a different geographic region. 

The CReSIS subset contains data from NASA’s Operation IceBridge campaigns conducted in Antarctica in 2009, covering predominantly polythermal regions that include both ice sheets and ice shelves. It comprises 18 missions, including 12 over West Antarctica and 6 over the Antarctic Peninsula, with 3 of the 18 missions flown at high altitudes. According to the benchmark dataset, we utilize 11 missions for training, 3 for validation, and 4 for testing. All splits include missions from both regions, while high-altitude missions are exclusively used for training. 

The FAU subset comprises radargrams acquired during multiple flights over three mountain glaciers. Specifically, three flights were conducted over the Perito Moreno Glacier in March and April, two over the Viedma Glacier during the same period, and two over the Gourdon Glacier in February. The Perito Moreno and Viedma Glaciers are both located in the Southern Patagonian IceField and mainly consist of temperate ice with high water content. This characteristic complicates the interpretation of the corresponding radargrams, as liquid water can scatter the radar signal, leading to ambiguity in the ice bottom position. In contrast, the Gourdon Glacier, located on James Ross Island  at the east side of the Antarctic Peninsula, consists primarily of polythermal ice. This makes the interpretation of the corresponding radargrams much clearer, as less noise is introduced into the radargram. For testing, we utilize the third flight of the Perito Moreno Glacier, the second flight over the Viedma Glacier, and the second flight over the Gourdon Glacier. The remaining flights are used for training and validation. 

The AWI subset includes radargrams from the Antarctic Peninsula and East Antarctica. Despite its broad physical coverage, the subset is characterized by a coarse horizontal resolution, which results in a limited number of annotated data points. In particular, the test set does not contain a single radargram segment with more than 2048 annotated traces, preventing a meaningful evaluation of our hooking mechanism, which utilizes a context width of 2048 pixels. Furthermore, due to the different processing schemes between the three collected flights, it is also not possible to rearrange the subset to isolate a single flight for testing. Consequently, we decided to exclude the AWI subset from our experiments.

A limitation of the standardized IceAnatomy data splits is their limited ability to assess model generalization to newly acquired campaign data. Although test and train sets consist of mutually exclusive flights, they are not region-, season-, or campaign-exclusive. As a result, models trained and evaluated under this protocol are not required to generalize to unseen campaigns, which constrains their practical applicability. Furthermore, this setup divides the available flights for each region in training and testing, substantially restricting the geospatial context for both. Consequently, the model has reduced capacity to learn meaningful geospatial relationships, and the evaluation does not fully reflect the potential of the \ac{geoconv}.
To address these issues, we introduce regionally exclusive splits to validate our proposed architectural enhancements in additional experiments. For the CReSIS dataset, all missions conducted in West Antarctica are used for training and validation, while missions from the Antarctic Peninsula are reserved for testing. In the FAU dataset, training and validation are performed using data from the Viedma and Gourdon Glaciers, with the Perito Moreno Glacier reserved for testing. To ensure generalization to new campaign data, we extend the FAU test set with newly acquired data from the Upsala Glacier in the Southern Patagonian Ice Field~\citep{Moritz}. Upsala Glacier is also a temperate mountain glacier and faces similar challenges to the Perito Moreno and the Viedma Glacier. Although its data were collected and processed using the same radar system and processing scheme, they were acquired during the spring season (October), in contrast to the Perito Moreno, Viedma, and Gourdon glaciers, where data acquisition primarily occurred between late summer and early autumn (February-April). As such, this dataset is both regionally and seasonally exclusive, providing an out-of-distribution test case representative of real-world deployment.

We note that cross-subset generalization between the FAU and CReSIS datasets is not considered, because differences in radar systems and processing pipelines result in a domain shift that is too large to permit meaningful evaluation.

\subsection{Experimental Protocol}
We conduct three sets of experiments to evaluate our proposed architecture. First, we compare our proposed approach against the baseline presented in Dreier et al.~\citep{me} using the standardized IceAnatomy data splits. For this, we train one omni model--defined as a model jointly trained on both the CReSIS and FAU subsets-- to enable comparison within the benchmark framework and to validate that our model can generalize to both subsets simultaneously.
Second, we assess the effectiveness of our new components using the regional exclusive splits defined in Section~\ref{sec:dataset}. Here, we train separate models for the FAU and CReSIS subsets to demonstrate the full potential of the proposed modifications under region-exclusive conditions and their capacity to specialize to distinct data sources.
Third, we evaluate the ability of our architecture to generalize to newly acquired campaign data. For this experiment, we evaluate the model trained on the regionally exclusive FAU split using unseen data from the Upsala Glacier.

To train our model, we rescale the radargrams to a consistent height of 1024 pixels and crop patches of $\mathrm{512\times1024}$ pixels. For our two new context modules, we require additional input data. First, the proposed hooking mechanism requires two additional radargrams of $\mathrm{1024\times1024}$ and $\mathrm{2048\times1024}$ resolution as spatial context. We center them on the original $\mathrm{512\times1024}$ crop and then resize to match the input resolution. Second, the geospatial context mechanism requires two additional radargrams in physical proximity to the initial one.  We choose these radargrams according to the rules defined in Section~\ref{meth:geocontext}, and set the look-up radius to $250\,\mathrm{m}$ for the CReSIS subset and $50\,\mathrm{m}$ for the FAU subset to account for the significantly different geospatial distances between traces in the two subsets.

As optimizer, we employ AdamW~\citep{adamw} with a batch size of 32 and a learning rate of 0.0005. We halve the learning rate if the ice-bottom \ac{map1} does not improve for 10 epochs, and we stop training after 35 consecutive epochs without improvement. Every epoch consists of 3000 samples, except for the experiments using only the FAU subset, where 500 samples are used due to the reduced dataset size. To prioritize deeper and more challenging ice bottoms during training, we weight the probability of choosing a sample from a respective flight or section by the sum of the ice surface and ice bottom picks, normalized by the total sum of ice surface and ice bottom picks over the entire dataset. Afterward, we randomly crop a sample from the chosen section. For augmentations, we employ random vertical flips with a probability of $\mathrm{p=0.5}$, gamma correction $\mathrm{p=0.2}$, random brightness and contrast adjustments with $\mathrm{p=0.2}$, and random noise with $\mathrm{p=0.2}$.

\section{Results}
\label{sec:results}

\begin{table*}[htbp]
       \begin{minipage}{\textwidth}
        \centering
        \caption{
        This table summarizes the quantitative comparison of our proposed model with the baseline on the standardized splits of the IceAnatomy dataset. Both approaches were trained on the FAU and CReSIS subsets simultaneously across five independent runs, and the reported values represent the mean and standard deviation. The ensembling model consists of the combination of the five trained models. A more detailed version of the results can be found in the Appendix \ref{appendix:breakdown} in Tab.~\ref{tab:omni:cresis:breakdown} and \ref{tab:omni:fau:breakdown}.}
        \label{tab:omni}
      \begin{tabular}{c|cccc|cccc}
            \toprule
            & \multicolumn{8}{c}{\textit{Ice Boundary Delineation}}  \\
            & \multicolumn{4}{c}{\textit{FAU}} & \multicolumn{4}{c}{\textit{CReSIS}}  \\
             \textit{Model} & \textit{MAE} $\downarrow$ & \textit{MME} $\downarrow$ & \textit{AP-1\%}$\uparrow$ & \textit{AP-5\%}$\uparrow$ & \textit{MAE} $\downarrow$ & \textit{MME} $\downarrow$ & \textit{AP-1\%}$\uparrow$ & \textit{AP-5\%}$\uparrow$\\
            \midrule
             Baseline  & $25.8\pm5.0$ & $19.1\pm4.0$ & $67.9\pm2.4$ & $91.8\pm1.6$ & $16.4\pm4.7$ & $84.5\pm22.7$ & $86.3\pm1.3$ & $93.0\pm1.1$ \\
             Ours  & $18.8\pm4.2$ & $13.7\pm3.4$ & $70.5\pm5.3$ & $94.0\pm1.1$ & $11.0\pm1.9$ & $56.9\pm9.2$ & $88.8\pm1.3$ & $95.3\pm1.0$ \\
             \midrule
             +Ensemble  & $\mathbf{15.3}$ & $\mathbf{10.9}$ & $\mathbf{73.4}$ & $\mathbf{95.2}$ & $\mathbf{8.3}$ & $\mathbf{43.4}$ & $\mathbf{91.0}$ & $\mathbf{96.3}$ \\
            \bottomrule
        \end{tabular}
         \end{minipage}
\end{table*}
As shown in Tab.~\ref{tab:omni}, our proposed additions to the omni model consistently outperform the baseline of Dreier et al.~\citep{me} across all metrics. The MME decreased by $5.4\,\mathrm{m}$ on the FAU-subset and by $27.6\,\mathrm{m}$ on the CReSIS subset, corresponding to a relative error reduction of $28.3\,\%$ and $32.6\,\%$, respectively. Incorporating ensembling further improves performance, yielding final MMEs of $10.9\,\mathrm{m}$ on the FAU subset and $43.4\,\mathrm{m}$ on the CReSIS subset. Similar improvements are observed for precision, with gains of $5.5$ and $4.7$ percentage points in \ac{map1}, and $3.4$ and $3.3$ percentage points in \ac{map5} compared to the baseline. Overall, these results suggest that the proposed approach improves both prediction accuracy and robustness across both subsets simultaneously.

\begin{figure*}[t]
					\centering
					\includegraphics[width=0.9\textwidth]{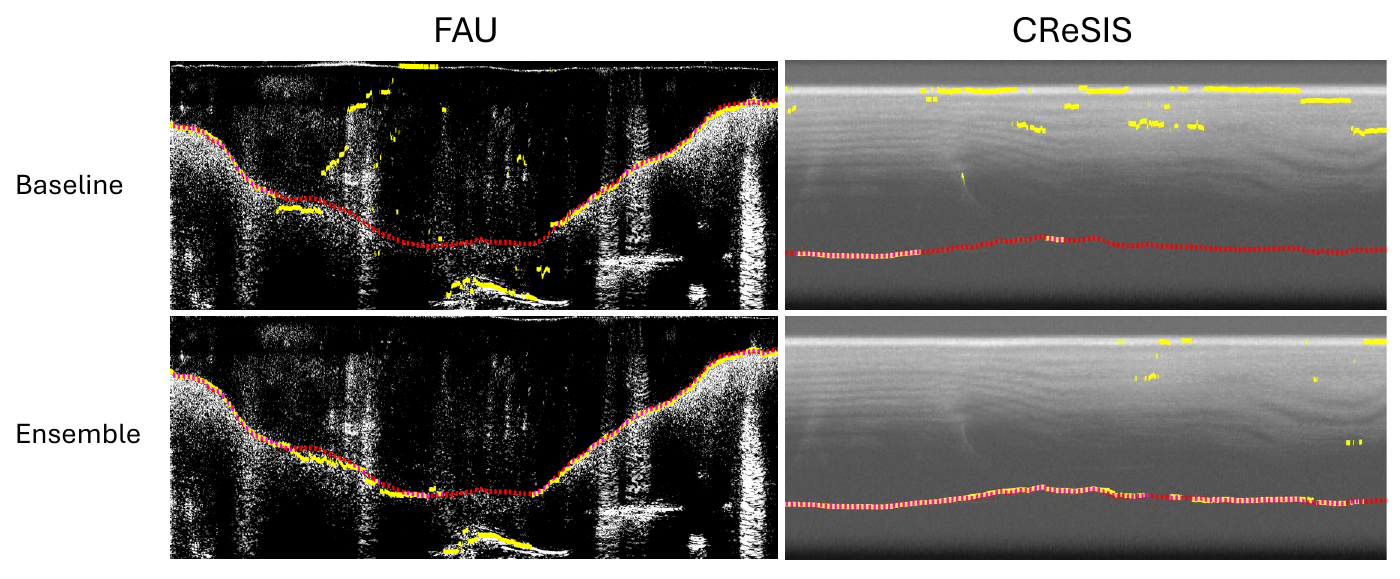}
					\caption{Qualitative results of the baseline and the ensemble model. Left: results from the third flight over the Perito Moreno Glacier portion of the FAU subset. Right: section from the TSK2-07 mission in the West Antarctica portion of the CReSIS subset. The \fboxsep=1pt\colorbox{yellow!100}{yellow} line indicates the ice bottom prediction, while the \fboxsep=1pt\colorbox{red!100}{red} dotted line indicates the ground truth.}
\label{fig:qual:omni}
\end{figure*}

These observations are also reflected in the qualitative results presented in Fig.~\ref{fig:qual:omni}. In the depicted radargrams from the Perito Moreno Glacier (FAU subset) and West Antarctica (CReSIS subset), the ensemble model remains more closely aligned with the ground truth in deeper ice layers. In contrast, the baseline model often confuses the ice surface with the bottom of the ice. Moreover, the ensemble model enhances the spatial consistency across consecutive predictions, as the frequency of large jumps in the predicted layer decreases. Nonetheless, both models still struggle in the deepest sections of the radargram, where visual features become unclear due to the weak return signal. 

\begin{table*}[htbp]
       \begin{minipage}{\textwidth}
        \centering
        \caption{Results on the IceAnatomy dataset using the newly defined region-exclusive splits. To reduce the influence of statistical outliers, each experiment is repeated five times, and the mean and standard deviation are reported. For the ensemble, results are reported for a single ensemble instance, as repeating the experiment five times would require training $5\times5$ individual models and incur substantial computational cost. A detailed breakdown of the results is provided in Appendix~\ref{appendix:breakdown}, Tables~\ref{tab:components:cresis:breakdown} and~\ref{tab:components:fau:breakdown}.}

        \label{tab:main_results}
      \begin{tabular}{c|cccc|cccc}
            \toprule
            & \multicolumn{8}{c}{\textit{Ice Boundary Delineation}}  \\
            & \multicolumn{4}{c}{\textit{FAU}} & \multicolumn{4}{c}{\textit{CReSIS}}  \\
             \textit{Model} & \textit{MAE} $\downarrow$ & \textit{MME} $\downarrow$ & \textit{AP-1\%}$\uparrow$ & \textit{AP-5\%}$\uparrow$ & \textit{MAE} $\downarrow$ & \textit{MME} $\downarrow$ & \textit{AP-1\%}$\uparrow$ & \textit{AP-5\%}$\uparrow$\\
            \midrule
             Baseline  & $50.9\pm8.1$ & $42.7\pm6.8$ & $48.5\pm0.7$ & $81.4\pm1.1$ & $15.4\pm1.4$ & $87.2\pm8.2$ & $82.3\pm1.2$ & $91.8\pm0.7$ \\
             +GeoConv  & $42.7\pm2.3$ & $35.8\pm1.9$ & $49.4\pm1.2$ & $83.7\pm0.5$ & $14.7\pm1.2$ & $83.4\pm6.2$ & $82.7\pm2.0$ & $92.5\pm0.7$ \\
             +Hooking & $36.7\pm5.0$ & $30.8\pm4.2$ & $50.8\pm1.9$ & $85.9\pm1.4$ & $13.5\pm1.0$ & $75.9\pm6.1$ & $84.2\pm2.1$ & $93.4\pm1.0$ \\
             \midrule
             +Ensemble & $\mathbf{26.1}$ & $\mathbf{21.9}$ & $\mathbf{55.2}$ & $\mathbf{89.0}$ & $\mathbf{11.1}$ & $\mathbf{61.4}$ & $\mathbf{87.5}$ & $\mathbf{94.8}$ \\
            \bottomrule
        \end{tabular}
         \end{minipage}
\end{table*}

An analysis of the region-specific splits in Tab.~\ref{tab:main_results} shows that each proposed component yields consistent performance improvements. On the FAU and CReSIS subsets, the introduction of the GeoConv reduces the MME by $6.9\,\mathrm{m}$ and $\mathrm{3.8}\,\mathrm{m}$, respectively. Incorporating the hooking mechanism yields further improvements of $\mathrm{5.0}\,\mathrm{m}$ on the FAU and $\mathrm{7.5}\,\mathrm{m}$ on the CReSIS subset. Finally, the addition of ensembling results in an MME of $21.9\,\mathrm{m}$ on the FAU subset, corresponding to a total error reduction of $\mathrm{48.7}\,\%$, and an MME of $\mathrm{61.4}$ on the CReSIS subset, corresponding to a total reduction of $\mathrm{29.6}\%$. Alongside these reductions, both \ac{map1} and \ac{map5} consistently increase with the introduction of each module. Compared to the baseline, the FAU subset exhibits gains of $\mathrm{6.7}$ and $\mathrm{7.6}$ percentage points in \ac{map1} and \ac{map5}, respectively, whereas the CReSIS subset shows improvements of $5.2$ and $3.0$ percentage points.

\begin{figure*}[t]
					\centering
					\includegraphics[width=0.9\textwidth]{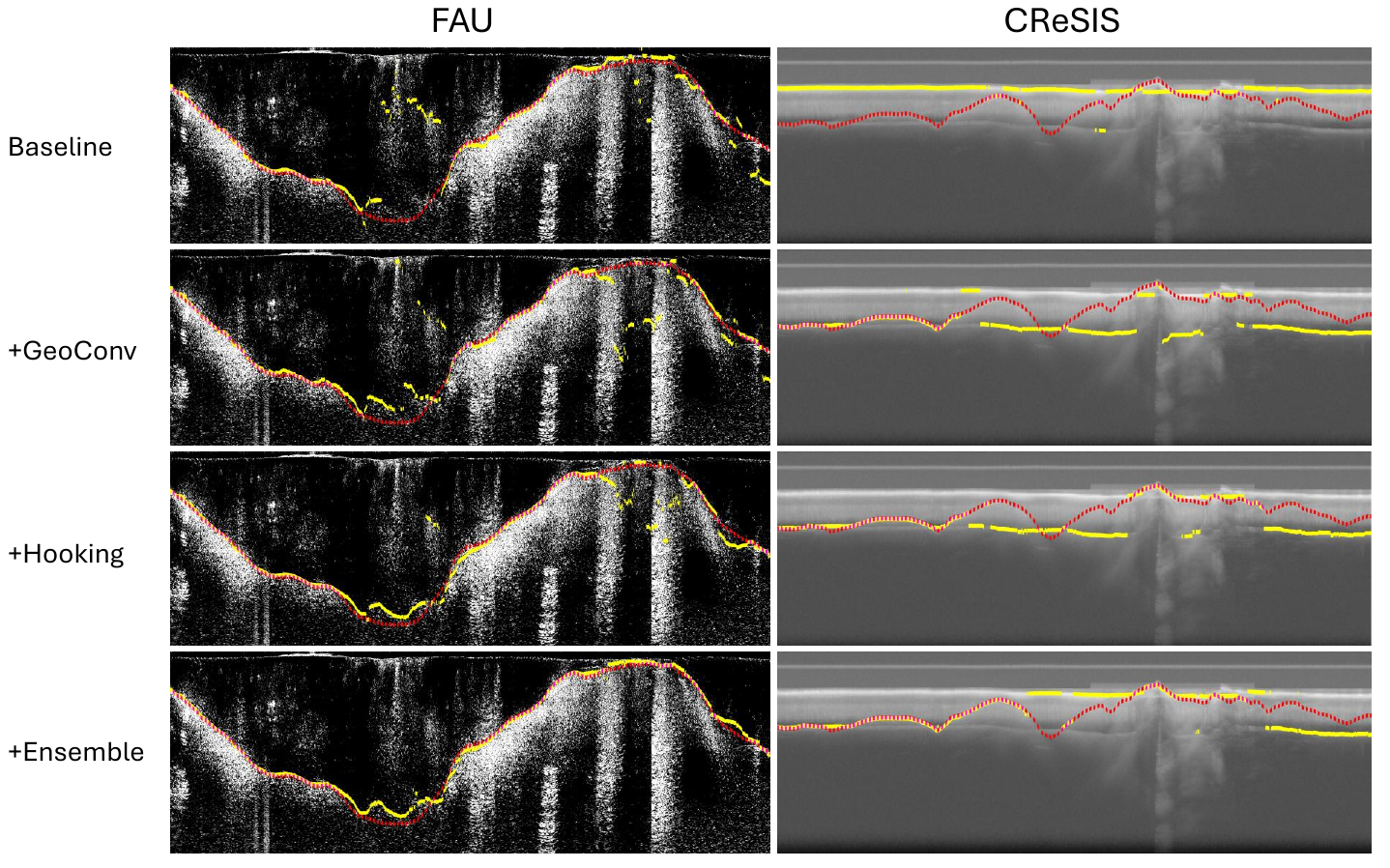}
					\caption{Qualitative comparison of model predictions as the proposed modifications are progressively incorporated. The left panels show radargrams from the first flight over the Perito Moreno Glacier, included in the FAU subset, while the right panels show a radargram from the PEN2-01 mission in the CReSIS subset. The \fboxsep=1pt\colorbox{yellow!100}{yellow} line denotes the predicted ice-bottom, whereas the \fboxsep=1pt\colorbox{red!100}{red} dotted line denotes the ground-truth annotation.}

\label{fig:qual:fau:big}
\end{figure*}

Figure~\ref{fig:qual:fau:big} visualizes the consistent qualitative improvements with each proposed component. 
In the FAU subset example, the baseline model struggles to accurately delineate the ice bottom in the deeper regions of the glacier. Incorporating the GeoConv module mitigates this issue, enabling the model to approximate the layer location through the geospatial context; however, the resulting predictions remain fragmented and exhibit notable discontinuities. The introduction of the hooking mechanism substantially improves spatial coherence, yielding more continuous predictions in deeper glacier sections, with only minor outliers near the surface. Finally, the ensemble model further stabilizes the results by reducing outliers via prediction averaging.
Similar trends are evident in the CReSIS example in Fig.~\ref{fig:qual:fau:big}. The baseline model fails to accurately delineate the ice boundary, whereas the inclusion of GeoConv improves predictions in the deeper regions of the ice. With the addition of the hooking mechanism, the predicted boundaries align more closely with the ground truth, particularly in the upper portions of the radargrams. The ensemble then further improves the predictions by reducing outliers and minimizing sensitivity to visual artifacts.

\begin{table}[htbp]
        \centering
        \caption{Quantitative evaluation of our subset model on radio-echo sounding data from the Upsala Glacier. Results are reported as the mean and standard deviation across five independent runs. The ensemble model combines the predictions of the five individually trained models.}
        \label{tab:upsala}
      \begin{tabular}{c|cccc}
            \toprule
             \textit{Region} & \textit{MAE} $\downarrow$ & \textit{MME} $\downarrow$ & \textit{AP-1\%}$\uparrow$ & \textit{AP-5\%}$\uparrow$ \\
            \midrule
            Baseline & $80.8\pm17.4$ & $67.8\pm14.6$ & $50.9\pm1.0$ & $80.9\pm0.6$ \\
            +GeoConv & $57.8\pm8.0$ & $48.5\pm6.7$ & $55.8\pm2.2$ & $86.2\pm1.0$  \\
            +Hooking & $51.5\pm13.9$ & $43.2\pm11.7$ & $52.7\pm4.3$ & $86.8\pm2.1$ \\
            \midrule
             +Ensemble  & $\mathbf{40.8}$ & $\mathbf{34.2}$ & $\mathbf{57.2}$ & $\mathbf{89.4}$ \\
            \bottomrule
        \end{tabular}
\end{table}

\begin{figure*}[t]
					\centering
					\includegraphics[width=0.8\textwidth]{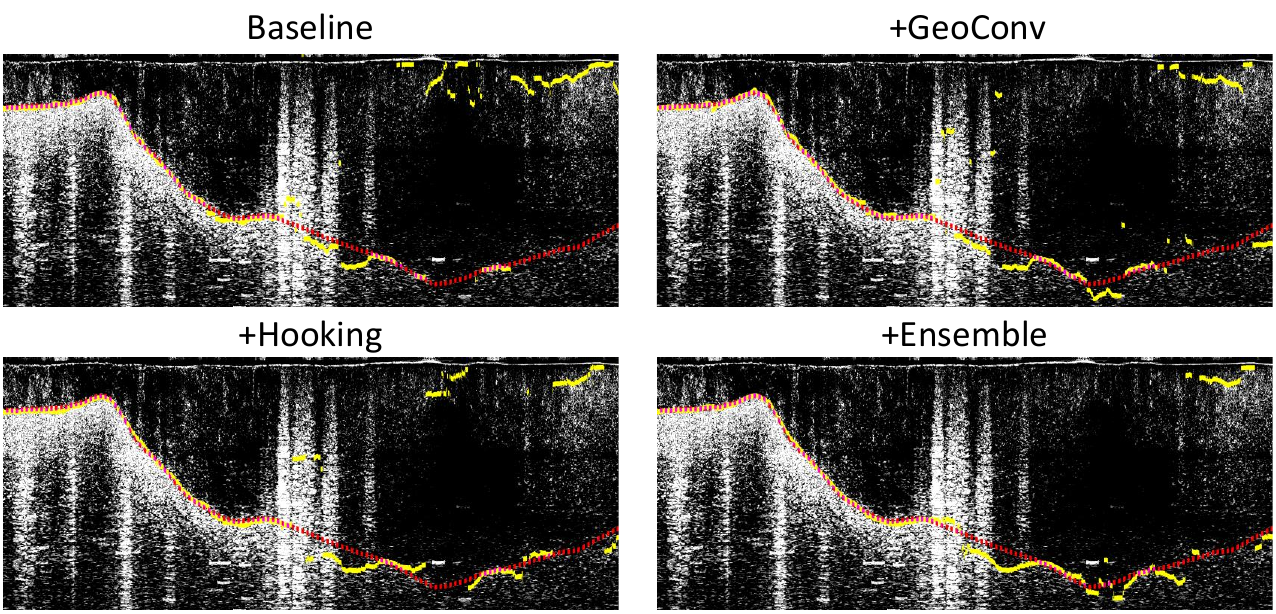}
					\caption{Qualitative comparison of the baseline and ensemble model predictions on the Upsala Glacier. The \fboxsep=1pt\colorbox{yellow!100}{yellow} line denotes the predicted ice-bottom, whereas the \fboxsep=1pt\colorbox{red!100}{red} dotted line denotes the ground-truth annotation.}
\label{fig:qual:upsala}
\end{figure*}

Examining the results of the final experiment on the Upsala Glacier in Tab.~\ref{tab:upsala}, we observe substantial performance improvements from each proposed component. Specifically, the MAE and MME decreased by $\mathrm{40.0}\,\mathrm{pixels}$ and $\mathrm{33.6}\,\mathrm{m}$, while the \ac{map1} and the \ac{map5} showed improvements of 6.3 and 8.5 percentage points. Despite these gains, the out-of-distribution Upsala experiment remains more challenging than simply isolating the Perito Moreno Glacier, as the new campaign introduces scenarios that diverge from prior conditions, such as the season or greater depths.
This is also indicated by the higher errors for the MAE and MME when compared to the regional-exclusive evaluation of the Perito Moreno Glacier. A qualitative comparison between the proposed modifications and the baseline in Fig.~\ref{fig:qual:upsala} indicates that these errors primarily arise from predictions of deep ice bottom layers. Nevertheless, the proposed approach helps to substantially reduce these errors.

\section{Discussion}
\label{sec:discussion}

Our experimental results demonstrate that our additions lead to substantial improvements on both the FAU subset and the CReSIS subset. Notably, the results with the omni model indicate that our modifications generalize to different types of radargrams and acquisition settings simultaneously, while achieving state-of-the-art results. A closer look at the results reveals that the proposed additions especially improve the detection of deeper ice bottom layers. We hypothesize that the extended context helps locate the ice boundaries when the visual features are insufficient or ambiguous. However, our current approach still suffers from occasional outlier predictions where the current context information seems to be insufficient to stabilize the prediction, resulting in discontinuities in the predicted layer. We hypothesize that such cases could be mitigated by incorporating additional structural contextual information from the predicted layer. For example, additional regularizers could favor smoother predictions or directly enforce structural coherence.

In the second set of experiments, we employed the newly proposed regional splits and subset-specific models. Our evaluation indicates that, on average, each additional component consistently improves the model's performance. In detail, the GeoConv provides broader contextual information via information from other flights that enables coarse-layer localization in deeper and ambiguous regions of the ice, while the hooking mechanism refines the coarse estimate via spatial cues from the surrounding context. The ensemble then further stabilizes the overall prediction by mitigating outliers through averaging over multiple model outputs. These findings suggest that the proposed components are especially beneficial for the localization of deep and ambiguous layers, as they introduce additional context clues that allow cross-validation between multiple patches and radargrams. Thus, the most challenging study sites with deep and ambiguous layers—namely the Perito Moreno Glacier in the FAU subset and West Antarctica in the CReSIS subset—exhibit the strongest performance increase compared to less complex sites such as the Viedma Glacier, Gourdon Glacier, or the Antarctic Peninsula. This trend is evident when analyzing the relative gains in the regional split and the general IceAnatomy split. For the FAU subset, the ensemble model achieves a higher relative gain on the region-specific split, which evaluates exclusively on the Perito Moreno Glacier, than on the general IceAnatomy split. In contrast, for the CReSIS subset, larger relative gains are observed under the general IceAnatomy split, which includes radargrams from West Antarctica, compared to the region-specific split restricted to the Antarctic Peninsula. However, performance across the CReSIS subset varies strongly among individual missions, reducing the regional influence on the overall evaluation. 

In the third and final experiment, we evaluate the FAU-subset-specific model on newly acquired data from the Upsala Glacier. Again, the proposed modifications outperform the baseline, setting state-of-the-art results. When directly comparing the results from the Perito Moreno and the Upsala Glacier, we notice that the \ac{mae} and \ac{mme} for the Upsala Glacier are substantially higher, despite both glaciers containing deep and structurally complex regions. This suggests that newly acquired campaign data may be more challenging than a region-exclusive subset such as Perito Moreno, as it may contain seasonal- or campaign-specific artifacts or a smaller and more restricted geospatial context. Nonetheless, our proposed modifications still improve the \ac{map1} and \ac{mme}, indicating that the proposed architectural additions can generalize effectively to newly acquired campaign data. Moreover, the relative performance gains are consistent with those observed for the Perito Moreno Glacier, further supporting the hypothesis that the proposed components are particularly beneficial for deep and structurally complex regions. However, the visual examination indicates that, despite noticeable improvements, the detection of very deep layers, especially those extending over large spatial extents, remains challenging. We hypothesize that these limitations stem from two main factors: first, deeper layers exhibit weaker return signals, which inherently complicates their detection, and second, the subset-specific model was trained exclusively on data from the Viedma and Gourdon Glaciers, which rarely exhibit such extreme extended depths, thereby introducing a bias toward shallower regions.

\section{Conclusion}
\label{sec:conclusion}
In this work, we extend a U-Net baseline by introducing two additional modules designed to incorporate contextual information. The first one, \ac{geoconv}, provides geospatial context to the model by processing radar traces according to their spatial positions. The second one is a new hooking module designed for radio-echo sounding data, which expands the model’s spatial view and enables greater spatial reasoning. 

Our evaluation demonstrates that both extensions substantially improve upon the initial baseline model. When combined with ensembling, our approach achieves a new state-of-the-art omni model performance on the IceAnatomy benchmark dataset, with errors of 10.9\,m for the FAU subset and 43.4\,m for the CReSIS subset. Compared to the baseline, this corresponds to a relative decrease in error of 42.9\,\% and 48.6\,\%, respectively. 
Furthermore, we propose two additional challenging experimental setups. In the first setup, the model is retrained for the FAU and CReSIS subsets individually; however, the data splits are rearranged to be geospatially exclusive. In the second setup, one of the models obtained from the first experiment is further evaluated on previously unseen campaign data from the Upsala glacier to assess its generalization capability to new campaign conditions. 
For both experiments, the proposed components lead to consistent performance improvements. In the first setup, the ensemble model achieves an \ac{mme} of 21.9\,m for the FAU subset and 61.4\,m for the CReSIS subset, corresponding to an error reduction of $\textrm{48.7\%}$ and $\textrm{29.6\%}$ compared to the baseline method. For the second experiment, the proposed approach achieves an overall error of 34.2\,m, corresponding to a relative improvement of 49.6\,\% compared to the baseline. These results highlight the importance of incorporating contextual information to accurately delineate ice boundaries.

{
\appendices%
\section{Runtime Optimizations}
\label{appendix:runtime}
Although the baseline model is relatively lightweight, the proposed modifications introduce computational overhead. This cost could be reduced in future work through several straightforward inference strategies. First, when sampling geospatially proximate regions, unevaluated sections could be prioritized to avoid redundant computations. Second, unlike during training, larger context windows do not need to be processed during inference once their final interaction with the target granularity has occurred. Furthermore, by restricting the \ac{geoconv} and the hooking module to the encoder, the decoder would not need to process additional context images during inference, thereby substantially limiting the computational overhead. Alternatively, the additional context outputs could be leveraged to further refine the final prediction. For example, multiple outputs corresponding to the same location could be weighted and combined to produce a single stable prediction.

\section{Additional results}
\label{appendix:breakdown}
To demonstrate the effects of the proposed modifications in various geographical settings, we evaluate each conducted mission and report the results. The performance of the omni model on the general IceAnatomy splits is summarized in Tabs. \ref{tab:omni:cresis:breakdown} and \ref{tab:omni:fau:breakdown}, while Tabs. \ref{tab:components:cresis:breakdown} and \ref{tab:components:fau:breakdown} present the results for region-exclusive splits using subset-specific models. From the results, we can see that the proposed ensemble model outperforms the baseline across all analyzed settings.

\begin{table}[!htbp]
    \centering
    \caption{Comparison of the proposed ensemble model and the baseline method across different geographical settings in the CReSIS subset. Both models are trained on the FAU and CReSIS subsets using the official train-test splits of the Ice-Anatomy benchmark dataset.}

    \label{tab:omni:cresis:breakdown}
      \begin{tabular}{c|cc|cc}
            \toprule
              Mission & \multicolumn{2}{c}{\textit{Baseline}} & \multicolumn{2}{c}{\textit{Ensemble}}  \\
              & \ac{mme} $\downarrow$ & \ac{map1} $\uparrow$ & \ac{mme} $\downarrow$ & \ac{map1} $\uparrow$ \\
             \midrule
             \multicolumn{5}{c}{\textit{West Antarctica}} \\
             TSK2-04 & $192.5$ & $68.7$ & $\textbf{103.8}$ & $\textbf{74.0}$ \\
             TSK2-07 & $373.6$  & $74.7$& $\textbf{146.3}$ & $\textbf{85.2}$ \\
            PIG4-01 & $60.1$ & $83.7$& $\textbf{21.1}$ &  $\textbf{93.6}$ \\
             PIG4-02 & $68.9$ & $86.6$ & $\textbf{57.0}$ & $\textbf{88.9}$ \\
            \midrule
                         \multicolumn{5}{c}{\textit{Antarctic Peninsula}} \\
             PEN4-01 & $34.5$ & $93.3$& $\textbf{14.8}$ & $\textbf{95.6}$ \\
             PEN5-01 & $43.0$ & $86.8$ & $\textbf{24.2}$& $\textbf{92.3}$ \\
             PEN5-02 & $49.5$ & $82.6$ & $\textbf{24.7}$ & $\textbf{89.9}$ \\

            \bottomrule
        \end{tabular}
\end{table}

\begin{table}[!htbp]
    \centering
    \caption{Comparison of the proposed modifications and the baseline method across different geographical settings in the FAU subset. The models are trained on both the FAU and CReSIS subsets using the official splits of the Ice-Anatomy benchmark dataset.}

    \label{tab:omni:fau:breakdown}
      \begin{tabular}{c|cc|cc}
            \toprule
              Mission & \multicolumn{2}{c}{\textit{Baseline}} & \multicolumn{2}{c}{\textit{Ensemble}}  \\
              & \ac{mme} $\downarrow$ & \ac{map1} $\uparrow$ & \ac{mme} $\downarrow$ & \ac{map1} $\uparrow$ \\
             \midrule
         Perito Moreno 3 & $46.2$ & $42.6$ & $\textbf{22.2}$ & $\textbf{51.9}$ \\
             Viedma 2 & $26.3$  & $58.0$& $\textbf{15.8}$ & $\textbf{64.4}$ \\
            Gourdon 2 & $5.0$ & $83.3$& $\textbf{3.8}$ &  $\textbf{86.8}$ \\
            \bottomrule
        \end{tabular}
\end{table}

\begin{table*}[!htbp]
    \centering
    \caption{Effect of the proposed modifications on the CReSIS subset model evaluated using region-exclusive splits across different flights. The best-performing configurations are highlighted in \textbf{bold}.}
    \label{tab:components:cresis:breakdown}
      \begin{tabular}{c|cc|cc|cc|cc}
            \toprule
              Mission & \multicolumn{2}{c}{\textit{Baseline}} & \multicolumn{2}{c}{\textit{+GeoConv}} & \multicolumn{2}{c}{\textit{+Hooking}}  & \multicolumn{2}{c}{\textit{+Ensemble}}   \\
              & \ac{mme} $\downarrow$ & \ac{map1} $\uparrow$ & \ac{mme} $\downarrow$ & \ac{map1} $\uparrow$ & \ac{mme} $\downarrow$ & \ac{map1} $\uparrow$  & \ac{mme} $\downarrow$ & \ac{map1} $\uparrow$  \\
             \midrule
             PEN4-01 & $36.7$ & $92.6$ & $33.7$ & $93.5$ & $27.1$ & $93.8$ & $\textbf{18.0}$ & $\textbf{94.6}$ \\
             PEN5-01 & $69.8$ & $82.3$ & $71.2$ & $81.3$ & $62.6$ & $82.8$ & $\textbf{50.4}$ & $\textbf{86.8}$ \\
             PEN5-02 & $90.6$ & $76.8$ & $93.1$ & $78.0 $ & $81.1$ & $80.2$ & $\textbf{68.0}$ & $\textbf{84.7}$ \\
             PEN3-01 & $56.7$ & $87.8$ & $53.8$ & $88.5$ & $45.8$ & $89.6$ & $\textbf{31.2}$ & $\textbf{94.0}$ \\
             PEN3-02 & $197.7$ & $65.0$ & $230.6$ & $60.0$ & $208.8$ & $64.0$ & $\textbf{186.9}$ & $\textbf{69.5}$\\
             PEN2-01 & $220.5$ & $57.8$ & $179.0$ & $60.7$ & $152.6$ & $67.3$ & $\textbf{133.1}$ & $\textbf{70.0}$\\
             PEN2-02 & $162.9$ & $70.3$ & $153.1$ & $71.7$ & $163.7$ & $72.3$ & $\textbf{152.2}$ & $\textbf{75.5}$\\
             PEN1-01 & $17.4$ & $97.1$ & $23.3$ & $96.0$ & $18.3$ & $96.4$ & $\textbf{13.0}$ & $\textbf{97.2}$\\
             PEN1-02 & $51.2$ & $91.2$ & $46.8$ & $92.6$ & $40.4$ & $93.1$ & $\textbf{36.4}$ & $\textbf{93.9}$\\
             LVISPEN-01 & $206.5$ & $58.0$ & $188.2$ & $56.8$ & $138.7$ & $67.0$ & $\textbf{86.3}$ & $\textbf{74.8}$\\
             LVISPEN-02 & $166.5$ & $62.5$ & $181.7$ & $58.6$ & $108.5$ & $73.3$ & $\textbf{62.9}$ & $\textbf{82.0}$\\
            \bottomrule
        \end{tabular}
\end{table*}

\begin{table*}[!htbp]
    \centering
   \caption{Effect of the proposed modifications on the FAU subset model evaluated using region-exclusive splits across different flights. The best-performing configurations are highlighted in \textbf{bold}.}

    \label{tab:components:fau:breakdown}
      \begin{tabular}{c|cc|cc|cc|cc}
            \toprule
              Mission & \multicolumn{2}{c}{\textit{Baseline}} & \multicolumn{2}{c}{\textit{+GeoConv}} & \multicolumn{2}{c}{\textit{+Hooking}}  & \multicolumn{2}{c}{\textit{+Ensemble}}   \\
              & \ac{mme} $\downarrow$ & \ac{map1} $\uparrow$ & \ac{mme} $\downarrow$ & \ac{map1} $\uparrow$ & \ac{mme} $\downarrow$ & \ac{map1} $\uparrow$  & \ac{mme} $\downarrow$ & \ac{map1} $\uparrow$  \\
             \midrule
             Perito Moreno 1a & $42.9$ & $49.6$ & $41.7$ & $50.0$ & $34.9$ & $50.8$ & $\textbf{27.6}$ & $\textbf{55.7}$ \\
             Perito Moreno 1b & $37.4$ & $51.4$ & $26.1$ & $51.2 $ & $20.5$ & $53.3$ & $\textbf{11.6}$ & $\textbf{60.8}$ \\
             Perito Moreno 2a & $54.2$ & $47.4$ & $42.9$ & $48.5 $ & $36.1$ & $50.5$ & $\textbf{23.5}$ & $\textbf{55.6}$ \\
             Perito Moreno 2b & $30.9$ & $45.1$ & $26.5$ & $45.6$ & $23.3$ & $48.4$ & $\textbf{20.5}$ & $\textbf{49.9}$ \\
             Perito Moreno 3 & $36.5$ & $50.2$ & $24.5$ & $53.3$ & $26.2$ & $52.9$ & $\textbf{13.0}$ & $\textbf{55.6}$\\
            \bottomrule
        \end{tabular}
\end{table*}

\section*{Code and Data Availability}
The code will be made publicly available on GitHub in the following repository \url{https://github.com/ki7077/ICIBS/tree/main}. The dataset can be accessed and downloaded from \url{https://zenodo.org/records/23368993} and \url{https://zenodo.org/records/23392580}.

\section*{Acknowledgments}
This research was funded by the Bayerisches Staatsministerium für Wissenschaft und Kunst within the Elite Network Bavaria with the Int. Doct. Program ``Measuring and Modelling Mountain Glaciers in a Changing Climate'' (IDP M3OCCA) as well as the German Research Foundation (DFG) project ``Large-scale Automatic Calving Front Segmentation and Frontal Ablation Analysis of Arctic Glaciers using Synthetic-Aperture Radar Image Sequences (LASSI)'' (grant number CH 2080/5-1 and SE 3091/4-1) and the project ``PAGE'' (grant number SE 3091/5-1) within the DFG Emmy-Noether-Programme.
The authors gratefully acknowledge the scientific support and HPC resources provided by the Erlangen National High Performance Computing Center (NHR@FAU) of the Friedrich-Alexander-Universität Erlangen-Nürnberg (FAU) under the NHR projects b110dc. NHR funding is provided by federal and Bavarian state authorities. NHR@FAU hardware is partially funded by the DFG – 440719683.
During the preparation of this work, AI technologies were used to assist in the writing process. Specifically, Grammarly (Grammarly, Inc., San Francisco, CA, USA) was used in order to check for grammar and style consistency and DeepL (DeepL SE, Cologne, Germany) and ChatGPT (GPT‐4) (OpenAI, San Francisco, CA, USA) were used in order to assist with rephrasing and improving readability. After using these tools, the manuscript was carefully reviewed and the content was edited as needed. No tools or services were used for content generation.

\section*{Author contribution}
\noindent
\textbf{Marcel Dreier}: Conceptualization, Methodology, Software, Project administration, Writing - Original draft preparation.
\textbf{Dakota Pyles}: Writing - review \& editing. 
\textbf{Nora Gourmelon}:  Writing - review \& editing.  
\textbf{Moritz Koch}: Data Collection, Writing - review \& editing. 
\textbf{Fei Wu}: Writing - review \& editing. 
\textbf{Thorsten Seehaus}: Supervision, Writing - review \& editing. 
\textbf{Matthias Braun}: Data Collection, Supervision, Writing – review \& editing. 
\textbf{Andreas Maier}: Supervision, Writing – review \& editing. 
\textbf{Vincent Christlein}: Supervision, Writing - review \& editing.

\bibliographystyle{IEEEtran}
\bibliography{IEEEabrv,bstctl,references}

\begin{thebibliography}{10}
\providecommand{\url}[1]{#1}
\csname url@samestyle\endcsname
\providecommand{\newblock}{\relax}
\providecommand{\bibinfo}[2]{#2}
\providecommand{\BIBentrySTDinterwordspacing}{\spaceskip=0pt\relax}
\providecommand{\BIBentryALTinterwordstretchfactor}{4}
\providecommand{\BIBentryALTinterwordspacing}{\spaceskip=\fontdimen2\font plus
\BIBentryALTinterwordstretchfactor\fontdimen3\font minus
  \fontdimen4\font\relax}
\providecommand{\BIBforeignlanguage}[2]{{%
\expandafter\ifx\csname l@#1\endcsname\relax
\typeout{** WARNING: IEEEtran.bst: No hyphenation pattern has been}%
\typeout{** loaded for the language `#1'. Using the pattern for}%
\typeout{** the default language instead.}%
\else
\language=\csname l@#1\endcsname
\fi
#2}}
\providecommand{\BIBdecl}{\relax}
\BIBdecl

\bibitem{rounce2023global}
D.~R. Rounce \emph{et~al.}, ``Global glacier change in the 21st century: Every
  increase in temperature matters,'' \emph{Science}, vol. 379, no. 6627, pp.
  78--83, 2023.

\bibitem{Intergovernmental_Panel_on_Climate_Change_IPCC2023-gl}
{Intergovernmental Panel on Climate Change (IPCC)},
  \emph{\BIBforeignlanguage{en}{Climate change 2022 -- impacts, adaptation and
  vulnerability}}.\hskip 1em plus 0.5em minus 0.4em\relax Cambridge, England:
  Cambridge University Press, Jun. 2023.

\bibitem{IPCC6}
K.~Calvin \emph{et~al.}, ``{IPCC}, 2023: Climate change 2023: Synthesis report.
  contribution of working groups i, {II} and {III} to the sixth assessment
  report of the intergovernmental panel on climate change [core writing team,
  h. lee and j. romero (eds.)]. {IPCC}, geneva, switzerland,''
  Intergovernmental Panel on Climate Change, Tech. Rep., Jul. 2023.

\bibitem{Hock_Maussion_Marzeion_Nowicki_2023}
R.~Hock, F.~Maussion, B.~Marzeion, and S.~Nowicki, ``What is the global glacier
  ice volume outside the ice sheets?'' \emph{Journal of Glaciology}, vol.~69,
  no. 273, p. 204–210, 2023.

\bibitem{essd-12-3039-2020}
\BIBentryALTinterwordspacing
E.~Welty \emph{et~al.}, ``Worldwide version-controlled database of glacier
  thickness observations,'' \emph{Earth System Science Data}, vol.~12, no.~4,
  pp. 3039--3055, 2020. [Online]. Available:
  \url{https://essd.copernicus.org/articles/12/3039/2020/}
\BIBentrySTDinterwordspacing

\bibitem{koch2025state}
M.~Koch \emph{et~al.}, ``The state and fate of glaciar perito moreno
  patagonia,'' \emph{Communications Earth \& Environment}, vol.~6, no.~1, p.
  572, 2025.

\bibitem{blindow2011performance}
N.~Blindow, C.~Salat, V.~Gundelach, U.~Buschmann, and W.~Kahnt, ``Performance
  and calibration of the helicoper gpr system bgr-p30,'' in \emph{2011 6th
  International Workshop on Advanced Ground Penetrating Radar (IWAGPR)}.\hskip
  1em plus 0.5em minus 0.4em\relax IEEE, 2011, pp. 1--5.

\bibitem{allen2012antarctic}
C.~Allen \emph{et~al.}, ``Antarctic ice depthsounding radar instrumentation for
  the nasa dc-8,'' \emph{IEEE Aerospace and Electronic Systems Magazine},
  vol.~27, no.~3, pp. 4--20, 2012.

\bibitem{sime2011automated}
L.~C. Sime, R.~C. Hindmarsh, and H.~Corr, ``Automated processing to derive dip
  angles of englacial radar reflectors in ice sheets,'' \emph{Journal of
  Glaciology}, vol.~57, no. 202, pp. 260--266, 2011.

\bibitem{tc-19-2159-2025}
\BIBentryALTinterwordspacing
H.~Moqadam and O.~Eisen, ``Review article: Feature tracing in radio-echo
  sounding products of terrestrial ice sheets and planetary bodies,'' \emph{The
  Cryosphere}, vol.~19, no.~6, pp. 2159--2196, 2025. [Online]. Available:
  \url{https://tc.copernicus.org/articles/19/2159/2025/}
\BIBentrySTDinterwordspacing

\bibitem{me}
\BIBentryALTinterwordspacing
M.~Dreier \emph{et~al.}, ``Iceanatomy: a benchmark dataset and methodology for
  automatic ice boundary extraction from radio-echo sounding data,'' \emph{The
  Cryosphere}, vol.~19, no.~11, pp. 5337--5359, 2025. [Online]. Available:
  \url{https://tc.copernicus.org/articles/19/5337/2025/}
\BIBentrySTDinterwordspacing

\bibitem{dong2021eisnet}
S.~Dong, X.~Tang, J.~Guo, L.~Fu, X.~Chen, and B.~Sun, ``Eisnet: Extracting
  bedrock and internal layers from radiostratigraphy of ice sheets with machine
  learning,'' \emph{IEEE Transactions on Geoscience and Remote Sensing},
  vol.~60, pp. 1--12, 2021.

\bibitem{cai2022accurate}
Y.~Cai, F.~Wan, S.~Hu, and S.~Lang, ``Accurate prediction of ice surface and
  bottom boundary based on multi-scale feature fusion network,'' \emph{Applied
  Intelligence}, vol.~52, no.~14, pp. 16\,370--16\,381, 2022.

\bibitem{kamangir2018deep}
H.~Kamangir, M.~Rahnemoonfar, D.~Dobbs, J.~Paden, and G.~Fox, ``Deep hybrid
  wavelet network for ice boundary detection in radra imagery,'' in
  \emph{IGARSS 2018-2018 IEEE International Geoscience and Remote Sensing
  Symposium}.\hskip 1em plus 0.5em minus 0.4em\relax IEEE, 2018, pp.
  3449--3452.

\bibitem{liu2022ice}
M.~Liu-Schiaffini, G.~Ng, C.~Grima, and D.~Young, ``Ice thickness from deep
  learning and conditional random fields: application to ice-penetrating radar
  data with radiometric validation,'' \emph{IEEE Transactions on Geoscience and
  Remote Sensing}, vol.~60, pp. 1--14, 2022.

\bibitem{jebeli2023tssa}
A.~Jebeli \emph{et~al.}, ``Tssa: two-step semi-supervised annotation for
  radargrams on the greenland ice sheet,'' in \emph{IGARSS 2023-2023 IEEE
  International Geoscience and Remote Sensing Symposium}.\hskip 1em plus 0.5em
  minus 0.4em\relax IEEE, 2023, pp. 56--59.

\bibitem{gifford2010automated}
C.~M. Gifford, G.~Finyom, M.~Jefferson, M.~Reid, E.~L. Akers, and A.~Agah,
  ``Automated polar ice thickness estimation from radar imagery,'' \emph{IEEE
  Transactions on Image Processing}, vol.~19, no.~9, pp. 2456--2469, 2010.

\bibitem{Mitchell.2013}
J.~E. Mitchell, D.~J. Crandall, G.~C. Fox, M.~Rahnemoonfar, and J.~D. Paden,
  ``A semi-automatic approach for estimating bedrock and surface layers from
  multichannel coherent radar depth sounder imagery,'' in \emph{Image and
  Signal Processing for Remote Sensing XIX}, L.~Bruzzone, Ed., vol. 8892.\hskip
  1em plus 0.5em minus 0.4em\relax SPIE, 2013, p. 88921E.

\bibitem{Rahnemoonfar.2017b}
M.~Rahnemoonfar, A.~A. Habashi, J.~Paden, and G.~C. Fox, ``Automatic ice
  thickness estimation in radar imagery based on charged particles concept,''
  in \emph{2017 IEEE International Geoscience and Remote Sensing Symposium
  (IGARSS)}, 2017, pp. 3743--3746.

\bibitem{Rahnemoonfar.2017}
M.~Rahnemoonfar, G.~C. Fox, M.~Yari, and J.~Paden, ``Automatic ice surface and
  bottom boundaries estimation in radar imagery based on level-set approach,''
  \emph{IEEE Transactions on Geoscience and Remote Sensing}, vol.~55, no.~9,
  pp. 5115--5122, 2017.

\bibitem{Crandall.2012}
D.~J. Crandall, G.~C. Fox, and J.~D. Paden, ``Layer-finding in radar echograms
  using probabilistic graphical models,'' in \emph{Proceedings of the 21st
  International Conference on Pattern Recognition (ICPR2012)}, 2012, pp.
  1530--1533.

\bibitem{Berger.2018}
V.~Berger, M.~Xu, S.~Chu, D.~Crandall, J.~Paden, and G.~C. Fox, ``Automated
  tracking of 2d and 3d ice radar imagery using viterbi and trw-s,'' in
  \emph{IGARSS 2018 - 2018 IEEE International Geoscience and Remote Sensing
  Symposium}, 2018, pp. 4162--4165.

\bibitem{Xu.2017}
M.~Xu, D.~J. Crandall, G.~C. Fox, and J.~D. Paden, ``Automatic estimation of
  ice bottom surfaces from radar imagery,'' in \emph{2017 IEEE International
  Conference on Image Processing (ICIP)}, 2017, pp. 340--344.

\bibitem{Lee.2014}
S.~Lee, J.~Mitchell, D.~J. Crandall, and G.~C. Fox, ``Estimating bedrock and
  surface layer boundaries and confidence intervals in ice sheet radar imagery
  using mcmc,'' in \emph{2014 IEEE International Conference on Image Processing
  (ICIP)}, 2014, pp. 111--115.

\bibitem{Ilisei.2014}
A.-M. Ilisei and L.~Bruzzone, ``A model-based technique for the automatic
  detection of earth continental ice subsurface targets in radar sounder
  data,'' \emph{IEEE GEOSCI REMOTE S}, vol.~11, no.~11, pp. 1911--1915, 2014.

\bibitem{Ilisei.2015}
------, ``A system for the automatic classification of ice sheet subsurface
  targets in radar sounder data,'' \emph{IEEE T GEOSCI REMOTE}, vol.~53, no.~6,
  pp. 3260--3277, 2015.

\bibitem{donini2021deep}
E.~Donini, F.~Bovolo, and L.~Bruzzone, ``A deep learning architecture for
  semantic segmentation of radar sounder data,'' \emph{IEEE Transactions on
  Geoscience and Remote Sensing}, vol.~60, pp. 1--14, 2021.

\bibitem{garcia2021automatic}
M.~H. Garc{\'\i}a, E.~Donini, and F.~Bovolo, ``Automatic segmentation of ice
  shelves with deep learning,'' in \emph{2021 IEEE International geoscience and
  remote sensing symposium IGARSS}.\hskip 1em plus 0.5em minus 0.4em\relax
  IEEE, 2021, pp. 4833--4836.

\bibitem{ghosh2022hybrid}
R.~Ghosh and F.~Bovolo, ``A hybrid cnn-transformer architecture for semantic
  segmentation of radar sounder data,'' in \emph{IGARSS 2022-2022 IEEE
  International Geoscience and Remote Sensing Symposium}.\hskip 1em plus 0.5em
  minus 0.4em\relax IEEE, 2022, pp. 1320--1323.

\bibitem{ghosh2022transsounder}
------, ``Transsounder: A hybrid transunet-transfuse architectural framework
  for semantic segmentation of radar sounder data,'' \emph{IEEE Transactions on
  Geoscience and Remote Sensing}, vol.~60, pp. 1--13, 2022.

\bibitem{ghosh2023enhanced}
------, ``An enhanced unsupervised feature learning framework for radar sounder
  signal segmentation,'' in \emph{IGARSS 2023-2023 IEEE International
  Geoscience and Remote Sensing Symposium}.\hskip 1em plus 0.5em minus
  0.4em\relax IEEE, 2023, pp. 6920--6923.

\bibitem{ghosh2023unsupervised}
------, ``An unsupervised framework for radar sounder signal segmentation based
  on enhanced self-supervised transformers,'' \emph{Authorea Preprints}, 2023.

\bibitem{ghosh2024urs}
------, ``Urs: an unsupervised radargram segmentation network based on
  self-supervised vit with contrastive feature learning framework,'' \emph{IEEE
  Journal of Selected Topics in Applied Earth Observations and Remote Sensing},
  vol.~17, pp. 15\,512--15\,524, 2024.

\bibitem{ronneberger2015u}
O.~Ronneberger, P.~Fischer, and T.~Brox, ``U-net: Convolutional networks for
  biomedical image segmentation,'' in \emph{International Conference on Medical
  image computing and computer-assisted intervention}.\hskip 1em plus 0.5em
  minus 0.4em\relax Springer, 2015, pp. 234--241.

\bibitem{cai2020end}
Y.~Cai, S.~Hu, S.~Lang, Y.~Guo, and J.~Liu, ``End-to-end classification network
  for ice sheet subsurface targets in radar imagery,'' \emph{Applied Sciences},
  vol.~10, no.~7, p. 2501, 2020.

\bibitem{VANRIJTHOVEN2021101890}
\BIBentryALTinterwordspacing
M.~{van Rijthoven}, M.~Balkenhol, K.~Siliņa, J.~{van der Laak}, and F.~Ciompi,
  ``Hooknet: Multi-resolution convolutional neural networks for semantic
  segmentation in histopathology whole-slide images,'' \emph{Medical Image
  Analysis}, vol.~68, p. 101890, 2021. [Online]. Available:
  \url{https://www.sciencedirect.com/science/article/pii/S1361841520302541}
\BIBentrySTDinterwordspacing

\bibitem{sneayknet}
E.~Schnider \emph{et~al.}, ``Improved distinct bone segmentation in upper-body
  ct through multi-resolution networks,'' \emph{International Journal of
  Computer Assisted Radiology and Surgery}, vol.~18, no.~11, pp. 2091--2099,
  2023.

\bibitem{amdhooknet}
F.~Wu \emph{et~al.}, ``Amd-hooknet for glacier front segmentation,'' \emph{IEEE
  Transactions on Geoscience and Remote Sensing}, vol.~61, pp. 1--12, 2023.

\bibitem{van2024multi}
M.~van Rijthoven \emph{et~al.}, ``Multi-resolution deep learning characterizes
  tertiary lymphoid structures and their prognostic relevance in solid
  tumors,'' \emph{Communications Medicine}, vol.~4, no.~1, p.~5, 2024.

\bibitem{wu2025lightweight}
F.~Wu \emph{et~al.}, ``Lightweight cross-attention-based hooknet for historical
  handwritten document layout analysis: F. wu et al.'' \emph{International
  Journal on Document Analysis and Recognition (IJDAR)}, vol.~28, no.~3, pp.
  409--427, 2025.

\bibitem{wu2024contextual}
------, ``Contextual hookformer for glacier calving front segmentation,''
  \emph{IEEE Transactions on Geoscience and Remote Sensing}, 2024.

\bibitem{11296938}
------, ``Amd-hooknet++: Evolution of amd-hooknet with hybrid cnn–transformer
  feature enhancement for glacier calving front segmentation,'' \emph{IEEE
  Transactions on Geoscience and Remote Sensing}, vol.~64, pp. 1--22, 2026.

\bibitem{aspp}
L.-C. Chen, G.~Papandreou, I.~Kokkinos, K.~Murphy, and A.~L. Yuille, ``Deeplab:
  Semantic image segmentation with deep convolutional nets, atrous convolution,
  and fully connected crfs,'' \emph{IEEE transactions on pattern analysis and
  machine intelligence}, vol.~40, no.~4, pp. 834--848, 2017.

\bibitem{pointconv}
W.~Wu, Z.~Qi, and L.~Fuxin, ``Pointconv: Deep convolutional networks on 3d
  point clouds,'' in \emph{Proceedings of the IEEE/CVF Conference on computer
  vision and pattern recognition}, 2019, pp. 9621--9630.

\bibitem{multi-annotatorGPR}
\BIBentryALTinterwordspacing
E.~S. Mannerfelt. The svalbard radar interpretation experiment. [Online].
  Available: \url{https://radar.mannerfelt.org/howto}
\BIBentrySTDinterwordspacing

\bibitem{Moritz}
\BIBentryALTinterwordspacing
M.~Koch \emph{et~al.}, ``Bedrock topography and ice thickness distribution of
  three major patagonian outlet glaciers unveiled by helicopter-borne ground
  penetrating radar,'' \emph{Earth System Science Data Discussions}, vol. 2026,
  pp. 1--29, 2026. [Online]. Available:
  \url{https://essd.copernicus.org/preprints/essd-2025-678/}
\BIBentrySTDinterwordspacing

\bibitem{adamw}
I.~Loshchilov and F.~Hutter, ``Decoupled weight decay regularization,''
  \emph{arXiv preprint arXiv:1711.05101}, 2017.

\end{thebibliography}

\end{document}